# Design, Modelling and Validation of a Novel Extra Slender Continuum Robot for In-situ Inspection and Repair in Aeroengine

Mingfeng Wang, *Member, IEEE,* Xin Dong, Weiming Ba, Abdelkhalick Mohammad, Dragos Axinte, Andy Norton

*Abstract* - **In-situ aeroengine maintenance works are highly beneficial as it can significantly reduce the current maintenance cycle which is extensive and costly due to the disassembly requirement of engines from aircrafts. However, navigating in/out via inspection ports and performing multi-axis movements with end-effectors in constrained environments (e.g. combustion chamber) are fairly challenging. A novel extra-slender (diameter-to-length ratio <0.02) dual-stage continuum robot (16 degree-of-freedom) is proposed to navigate in/out confined environments and perform required configuration shapes for further repair operations. Firstly, the robot design presents several innovative mechatronic solutions: (i) dual-stage tendon-driven structure with bevelled disks to perform required shapes and to provide selective stiffness for carrying high payloads; (ii) various rigid-compliant combined joints to enable different flexibility and stiffness in each stage; (iii) three commanding cables for each 2-DoF section to minimise the number of actuators with precise actuations. Secondly, a segment-scaled piecewise-constant-curvature-theory based kinematic model and a Kirchhoff-elastic-rod-theory based static model are established by considering the applied forces/moments (friction, actuation, gravity and external load), where the friction coefficient is modelled as a function of bending angle. Finally, experiments were carried out to validate the proposed static modelling and to evaluate the robot capabilities of performing the predefined shape and stiffness.**

## I. Introduction

Continuum robots, unlike those based on conventional rigid-links, e.g. either 6-DoF (Degree of Freedom) industrial [1] or hyper-redundant [2] designs, are composed of multiple small-diameter bendable sections, which are commonly actuated by tendons/air from their base [3]. Each of these sections can form a compliant circular-like shape with the utilisation of flexible materials as backbones. With these features, continuum robots are able to access and operate in confined spaces for which a wide range of practical applications have been presented in the literature, such as minimally invasive surgery [4], [5], nuclear facility maintenance [6], aircraft assembly [7] ,[8], manipulation [9], locomotion [10] and so on. Recently, some successful researches have been reported on the use of continuum robots for in-situ engine repair which enable a prompt response to the unplanned inspection and repair of aeroengines [11], [12]; this not only eliminates significant disassembly/assembly costs but also the penalties for non-operational time.

However, challenges still remain in the aspects of both design and modelling of the extra slender continuum robots for industrial applications (e.g. inspection and repair deep inside of jet engines), especially when requiring the hyper-redundant arm to be built with a small diameter (e.g. <15 mm), long length (e.g. ＞500 mm) and an appropriate stiffness for carrying miniaturised industrial end-effectors. According to the approaches of actuations, most of the continuum robots can be divided into two categories: pneumatic- and tendon-driven. As a typical example of pneumatic driven ones, FESTO elephant truck [13] was developed for safe human-robot cooperation on production lines, which is composed of a 1.2-meter 9-DoF arm and made of soft materials. However, the large diameter of the arm (114 mm) makes it unable to access into confined spaces for conducting tasks, which is a generic challenge for constructing small diameter continuum robots with pneumatic actuation. Further, tendon-driven actuation is another widely used approach. A number of rigid links can be serially connected to form a hyper-redundant robot arm (generally ＞1.5-meter long) and each of rigid links is driven by three/four steel cables [2], [7]. Compared with the pneumatic actuated arms, these tendon-driven ones can carry a significantly heavier payload. However, if a rigid link structure is adopted, this not only poses challenges to miniaturise the robot but also limits the accessibility in confined spaces, especially at sharp corners as all of them present a strict range of pivoting angles. Thus, to improve accessibility, flexible backbones are employed to enable miniaturisation and high bending angles [5]-[7]. For example, in the medical field, flexible backbone continuum robots are widely employed for minimally invasive surgery [5]. However, the length of these robots (e.g. < 300mm [5]) is generally much shorter than the rigid-link hyper-redundant ones (e.g. >1400mm [2]) while the payloads are very limited (e.g. <150g [5]). To the best of authors' knowledge, no systems exist that can meet all of the requirements (i.e. small diameter, long length, appropriate stiffness and hyper-redundant DoFs) simultaneously to enable demanding industrial applications such as in-situ repair interventions.

Apart from the design challenge, modelling (e.g. kinematics and statics) is another key aspect for realising continuum robots to be utilised in industrial applications. In terms of kinematic

This work was supported by the Aerospace Technology Institute (UK) under Grant Agreement No. 102360 (FLARE) and the Industrial Strategy Challenge Fund (UK) managed by EPSRC under Grant Agreement No. EP/R026084/1 (RAIN). *(Corresponding author: Xin Dong).*
M. Wang, X. Dong, W. Ba, A. Mohammad, D. Axinte are with the Rolls-Royce UTC in Manufacturing and On-Wing Technology, University of Nottingham, Nottingham, NG8 1BB, UK *(e-mail: {Mingfeng.Wang, Xin.Dong, Weiming.Ba, Abd.Mohammad, Dragos.Axinte} @nottingham.ac.uk).*

A. Norton is with Rolls-Royce plc, Derby, DE24 8BJ, UK *(e-mail: Andy.Norton@rolls-royce.com).*

modelling, most of the existing approaches are derived by assuming each section of the robot bends as a constant curvature [13], [14], which can significantly simplify the model. However, in fact, several practical facts (e.g. friction between the cables and guiding holes and gravity of the sections) affect the actual shape of sections of the arm, which can lead to significant errors with increase of length of the compliant joints with the widely used assumption of constant curvature [16]. Another approach for continuum robot kinematic modelling is variable curvature model, which involves static modelling. Generally, principle of virtual work [17] and Cosserat rod theory [18] are used for studying the section shape. Those methods consider the effects of gravity, friction, torsion, and external loading, which have been validated to a continuum robot with pure flexible backbone [16]. However, a novel structure of a continuum robot arm is proposed in this paper, which is a hybrid mechanism composed of a combination of complaints and rigid joints along the entire arm. Hence, a deep understanding of the fundamental kinematics and statics of the new extra slender continuum robots is needed.

In this paper, to develop a novel in-situ aeroengine maintenance technology, the technical requirements for performing invasive repair of aeroengines via borescope inspection ports and the technical challenges to be tackled by the proposed extra slender continuum robot are specified in Section II. In Section III, a novel design solution of a continuum robot is proposed by employing a combination of compliant and rigid joints, which can meet all the requirements of small diameter, long length, appropriate stiffness and hyper-redundant DoFs simultaneously. Further, different from conventional kinematics that employ piecewise constant-curvature theory (PCCT) to each section, a new segment-PCCT based kinematic model is established. After that, a Kirchhoff elastic rod theory is employed to develop the static model (linked with the segment-based kinematic model) of the new extra slender continuum arm to enable precise control of the robot arm. In Section IV, a prototype of the proposed continuum robot is built and a series of validation tests of this complex mechatronic system is performed to check the key characteristics to the targeted demonstration. Finally, conclusions are presented in Section V.

## II. REQUIREMENTS IDENTIFICATION

Delamination of combustor liner Thermal Barrier Coatings (TBCs) during aeroengine running is possible, which in turn exposes the metallic substrate to the harsh environments during engine running. This can lead to oxidation and material degradation. Whilst not safety critical, this can involve costly inspections or early engine removals. Engine uptime can be improved for the airliners through the use of advanced in-service component repairs. In this paper, a novel in-situ aeroengine repair technology by the deployment of miniature flame spray gun into combustion chamber via borescope ports using slender continuum robot is proposed for which the conceptual design is shown in Fig. 1. Moreover, the system needs to be capable of carrying and positioning the minimised flame spay unit inside the combustion chamber, so that it may respray patches of worn or damaged TBC. Beyond the positional task, the continuum robot needs to be capable of delivering the necessary oxyacetylene gas and the powder materials required for coating repair to the spray unit via its central working channel.

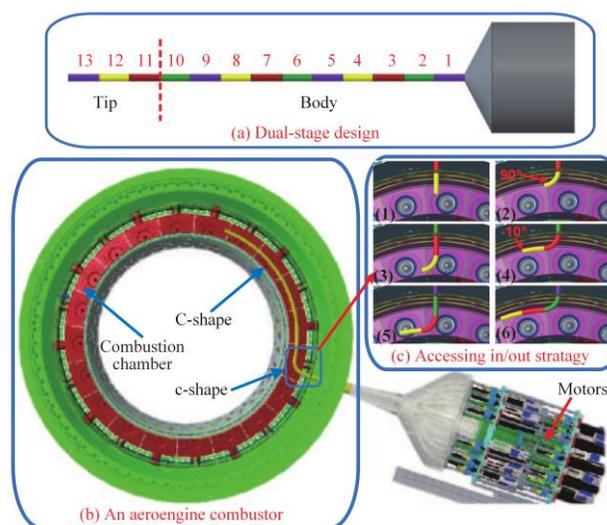

Fig. 1. A graphic representation of the conceptual design of the proposed dual-stage extra slender continuum robot: (a) a dual-stage design; (b) an aeroengine combustor; (c) accessing in/out strategy.

Therefore, the technical requirements of the proposed extra slender continuum robot for performing in-situ inspection and repair via borescope ports can be specified as follows:

• Arm length: no less than 700 mm for covering the overall internal area of the combustor by accessing the different inspection holes (Fig. 1(b));

• Arm diameter: no more than 15 mm in order to access into the combustor via the inspection holes (i.e. ports) included in the combustor casing;

• Section length: to access into the combustor without collision with the internal surface, each of the section length needs to be less than 60 mm (Fig. 1(c));

• Section number: no less than 11 sections, which is determined by the requirements of the arm and section lengths to access particular features within the aeroengine;

• Payload: minimum payload of 125g at the tip to ensure that the appropriate end-effectors (e.g. inspection and repair tools) can be adopted;

Overall, the analysis of tehcnical requirements led to the conclusion that to conduct the in-situ repair of aeroengine combustion chambers (e.g. redeposition of TBCs) tasks, an extra slender continuum robot with a small diameter (<15 mm), a long length (>700 mm) and an appropriate stiffness for carrying miniaturized industrial end-effectors needs to be developed, which could be considered as a big challenge according to the literature; note that most of the existing continuum robots are structured with no more than three sections for minimally invasive surgery. Further, the modelling of the new design of continuum arm is another challenge due to this unique dual (i.e. rigid and compliant joint) structure, while most of the existing ones assume that each of the sections bends as a circular arc and both the gravity and the friction are neglected. However, this assumption cannot be applied to the extra slender continuum robot as it will result in significant positioning errors.

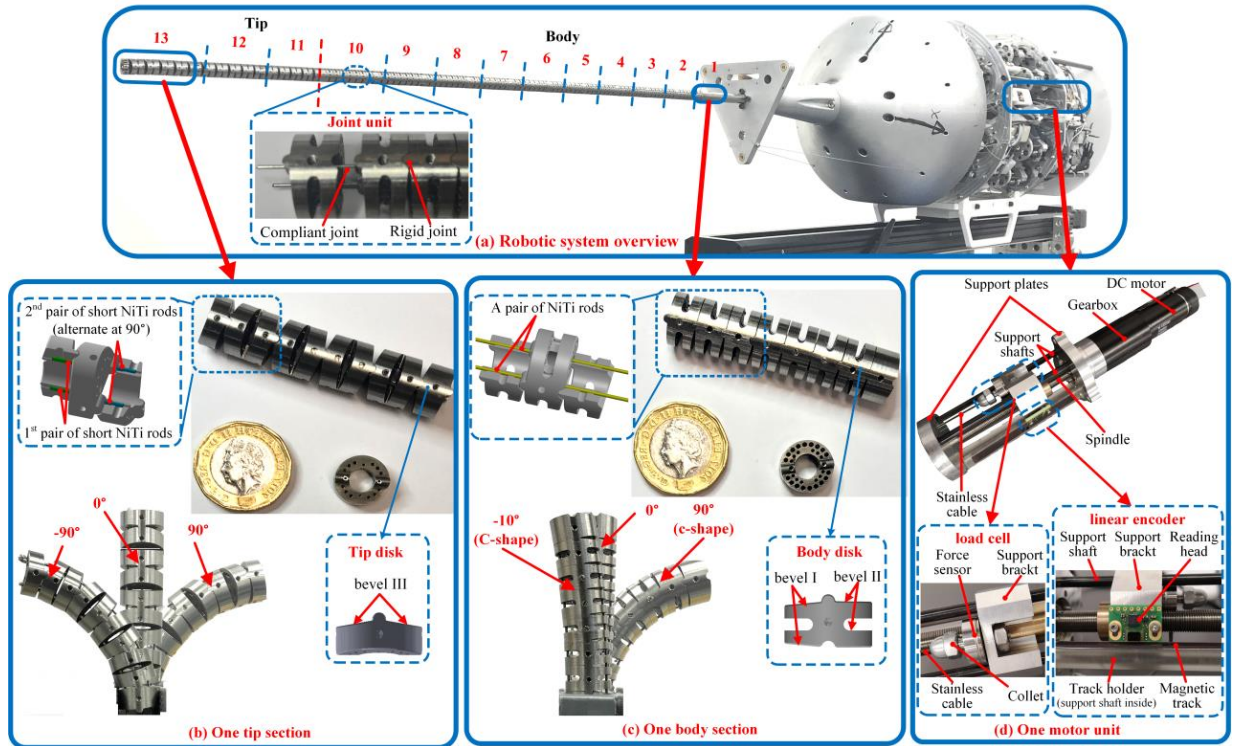

Fig. 2. Design of the proposed continuum robotic system: (a) Overview of the robotic system: sections 1-10 (1 DoF each) representing the Body of the robot; sections 11-13 (2 DoFs each) representing the Tip of the robot; (b) detail of one tip section; (c) detail of one body section and (d) detail of one motor unit.

## III. Solutions: Design and Modelling

In this section, by using a novel combination of compliant and rigid joints, the extra slender (diameter of 12.7 mm and length of 715 mm accounting for diameter-to-length ratio <0.02) and fully actuated continuum robot arm (16 DoFs) is presented, which is capable of carrying miniaturised industrial end-effectors. Then, coordinate systems and modelling assumptions are established for kinematic and static analyses of the proposed continuum robot arm. Finally, the piecewise constant-curvature theory and Kirchhoff elastic rod theory are respectively applied to establish the kinematic and static models of the proposed novel design for the first time, which forms the basis for mechanical design, actuation specification (i.e. tendon thickness and force) and controller development.

### A. Robotic System Design

The developed robotic system, as shown in Fig. 2, is composed of a continuum robot arm and a tendon-driven actuation module for position and force control of the first and the main specifications listed in TABLE I.

In the continuum robot arm, there are two stages as shown in Fig. 2. In the first stage (tip sections), there are three sections and each section is structured with 2 DoFs, which enables 6-DoF movements of this stage for maintenance operation in confined spaces. The second stage (body sections) is composed of ten 1-DoF sections, which is able to deliver the tip sections to navigate in/out the aeroengine combustor, in form of a torus shape, via small inspection ports. Thus, the entire continuum robot is structured with 16 DoFs allocated to 13 sections.

Regarding the robot arm design, a novel solution of a twin-

TABLE I
MAIN SPECIFICATION OF THE CONTINUUM ROBOT

| Continuum Robot | Stage | |
|---|---|---|
| | Body (10 sections) | Tip (3 sections) |
| DoF | 10 (1 per section) | 6 (2 per section) |
| Mass | 141.17 g | 51.99 g |
| Length | 550 mm | 165 mm |
| Diameter-to-length ratio | < 0.02 | |
| Inner-to-outer diameter ratio | > 0.5 | |
| Bending capability[a] | -10° to +90° | -90° to +90° |
| Disk | 80 (8 per section) | 30 (10 per section) |
| Cable/Motor | 20 (2 per section) | 9 (3 per section) |
| Backbone | A pair of NiTi rods (ø 0.6 mm)[b] | A NiTi rod twin-pivot (ø 0.6 mm)[c] |
| Driven tendon | Stainless steel cables (ø 0.90 mm) | Stainless steel cables (ø 0.75 mm) |

a. Bending capability in each section;
b. A pair of NiTi rods run through all body sections;
c. A pair of short NiTi rods is allocated between two adjacent disks.

pivot revolute joint unit is proposed by utilising the combination of compliant (NiTi rods) and rigid (disks) joints (see an exploded view in Fig. 2(a)), which can ensure the high flexibility and enhance torsional stability. Specifically, in each body section, there are eight revolute joints constrained by a pair of superelastic NiTi rods (Euroflex®) as compliant joints, which are applied through the entire length of the second stage. Hence, each section performs as a 1-DoF continuum structure (see Fig. 2(c)). In the tip sections, a solution of the twin-pivot compliant joints (short NiTi rods) is adopted, which alternate at 90° to account for two DoFs in each section (see Fig. 2(b)).

In addition, a bevelled disk design with specific angled surfaces is applied in the continuum robot arm. In each disk of the body sections (see Fig. 2(c)), there are two sets of bevels (i.e. bevels I and II respectively are 1.25° and 11.25°). The angled surfaces are determined by the maximum bending angle of each body section when performing navigation in/out at the port (c-shape in Fig. 1) and navigation along the middle line of the combustor chamber (C-shape in Fig. 1). In each disk of the tip sections (see Fig. 2(b)), the slopes of four bevels (i.e. bevels III) on both sides are the same while two bevels on one side alternate at 90° relative to the other side.

To achieve the high payload capability at the tip and compensate the self-weight, a tendon-based and extrinsically actuated approach is applied, where a pair of and a group of three cables are respectively allocated to each body and tip sections. Furthermore, to prevent the backlash and slack, all cables are fully actuated which means totally 29 cable/motors (Orminston®/Maxon®) are employed. In each motor unit (seeFig. 2(d)), a magnetic linear encoder (Renishaw®) and a load cell (Omega®) are integrated to provide the position (i.e. displacement of the cable) and force feedback (i.e. proximal tension of the cable) for further advanced control strategy.

### B. Coordinate Systems and Modelling Assumptions

The proposed continuum robot is divided into two stages (i.e. body and tip), as shown in Fig. 3, where the tip (sections 11 to 13) and body (sections 1 to 10) can be modelled as a spatial and a planar continuum manipulators, respectively. A tip-following navigation algorithm [19] is applied to generate the path for the entire robot which allows the user to navigate the tip through a restrictive environment with the entire body following the tip's path. Once in the desired position, in our application, only the tip sections are used for performing maintenance tasks whilst the rest body sections are rigidised by interlocking on the bevelled surfaces, i.e. 90° (c-shape) or -10° (C-shape) in Fig. 2(c), which means the continuum robot is in an optimal position for maintenance operation with the positions of sections 1-10 fixed.

*1) Coordinate Systems:* To implement the kinematic and static analysis, the coordinate frames in terms of world $\{O\}$, section $\{O_i\}$, disk $\{O_{i,j}\}$ and end-effector $\{O_{14}\}$ frames are established (see Fig. 3). The general kinematic representation of configuration space (i.e. arc parameters in terms of the curvature $\kappa$, the rotational angle $\phi$ and the arc length $\ell$) is illustrated referring in the world frame $\{O\}$ in Fig. 3, and the arc geometry relationships can be presented as

$$\begin{cases} \theta = \kappa \cdot \ell \\ r = 1/\kappa \end{cases} \quad (1)$$

where $\theta$ is the bending angle and $r$ is the bending radius. Specifically, the $+z-$ axes are considered to be always tangent to the arc. Furthermore, the task space (i.e. the position and orientation of the end-effector) is represented by the position of the end-effector centre $O_{14} = [x_{14}, y_{14}, z_{14}]^T$ and the rotation matrix of the frame $\{O_{14}\}$: $^0R_{14} \in \Re^{3\times 3}$.

*2) Modelling Assumptions:* The following assumptions have been made for the kinematic and static models of the proposed extra slender continuum robot.

*A1* The pair or twin-pivot superelastic NiTi rods are considered to perform as a planar elastic rod which is thin and inextensible. This implies that the backbones in the body sections and each segment of tip sections are confined to a planar workspace without shear and extension deformations;

*A2* The disks are rigid and the friction between the cables and routing holes in the disks should not be neglected. In this paper, it is assumed that the static and the sliding friction coefficients are approximately equal. Note that the angle between the cable and its routing holes varies during the robot bending and this implies that the friction coefficient is a function of the bending angle rather than a constant value;

*A3* The locations of the cables within the cross-section of the robot are assumed not to change during the deformation, as the holes in the disks are with tight tolerances;

*A4* The continuum robot is always manipulated with static equilibrium.

### C. Kinematics Analysis

In this subsection, the piecewise constant-curvature theory (PCCT) [13] is utilised to formulate the kinematics of the continuum robot, which can provide an efficient way to describe the motion. Different from previous section-based PCCT methods, a segment-based one is proposed here, where each section is decomposed into multiple segments which are units of curvature and the different bending angles in segments are calculated based on the static model. By concatenating these segment curvatures, the kinematic model can describe the posture of the robot even if it presents uneven curvature.

Based on the PCCT, the continuum robot kinematics are generally decomposed into two submappings. One is between joint or actuator space (e.g. length of cable) and configuration space (arc parameters $q = [\kappa, \phi, \ell]^T$), while the other is between this configuration space and task space (i.e. position and pose of the end-effector). Furthermore, the former is a robot-specific mapping which means the kinematics varies with different actuation approaches (e.g. tendon-driven), while the latter is a robot-independent mapping which is general and applies to each independent actuated segment.

*1) Configuration-Cable Kinematics:* Since cable-driven is chosen as the actuation approach (i.e. the arc is shaped by driving cables) in the proposed continuum robot, the purpose of configuration-cable kinematic analysis is then to derive the equations that describe the lengths of the cables relative to the given bending arc configuration in each segment. As mentioned in Section III.A, two cables and a group of three cables are respectively utilised to actuate each body (1 to 10) and tip (11 to 13) sections; hence, the lengths of all cables can be written in the forms of $L = [l_1, \ldots, l_{29}]^T$.

Noting that the cables are allocated to disk holes with different pitch circle diameters (PCDs: $D_1$, $D_2$ and $D_3$) and phase angles (e.g. $H_{i,j,k}$ and $\varphi_{i,j,k}$ respectively indicate the routing hole and phase angle of the $k^{th}$ cable at the $j^{th}$ disk of $i^{th}$ section), as shown in Fig. 3, the PCD projection (red dashed lines) of $H_{i,j,k}$ on the bending plane can be then expressed as

$$\begin{cases} |D_1 \cdot \sin\varphi_{i,j,k}|, \text{when } i = 1, \ldots, 9; k = 1, \ldots, 18 \\ |D_2 \cdot \sin\varphi_{i,j,k}|, \text{when } i = 1, \ldots, 10; k = 19, \ldots, 29 \\ |D_3 \cdot \sin\varphi_{i,j,k}|, \text{when } i = 11, 12, 13; k = 21, \ldots, 29 \end{cases} \quad (2)$$

where the cables ($k = 1, \ldots, 18$) for actuating body sections 1 to 9 pass through the body disk holes with $D_1$ (see Fig. 3(a)), the pair of cables ($k = 19, 20$) for section 10 are through the body disk holes with $D_2$ (see Fig. 3(a)), while the ones ( $k = $ 21,…,29) for tip sections 11 to 13 first route through body disk holes with $D_2$ and then tip disk holes with $D_3$ (see Fig. 3(b)).

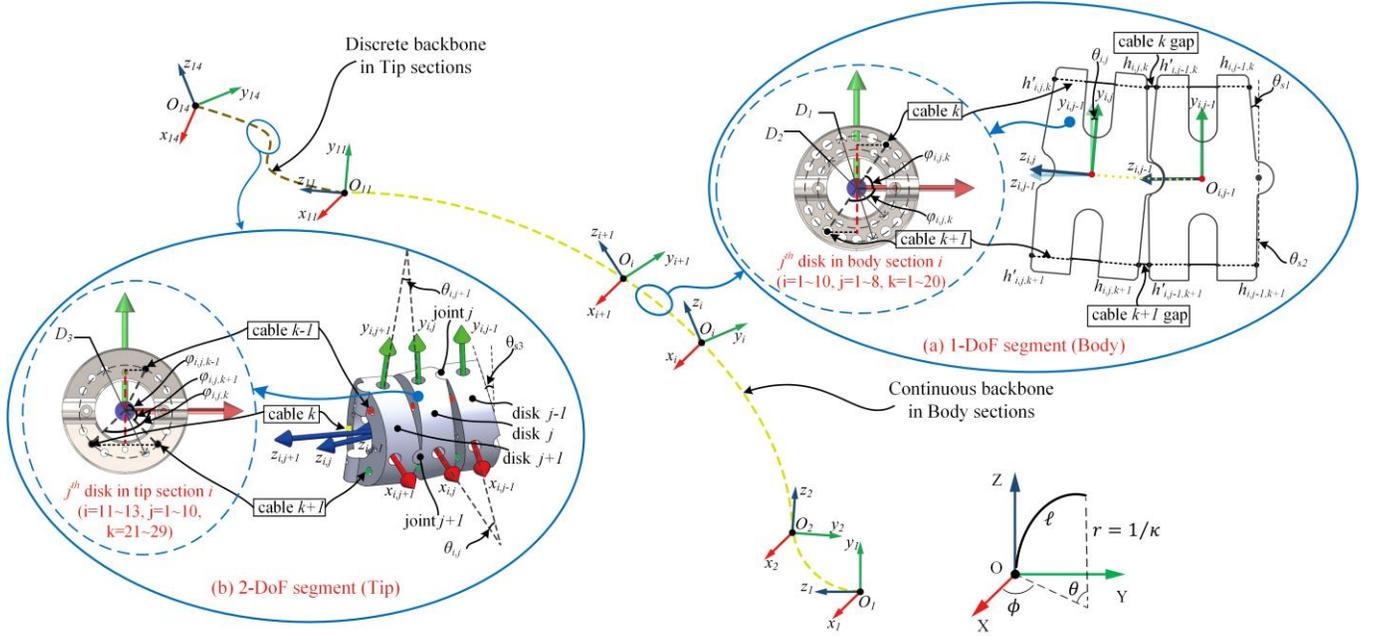

Fig. 3. Schematic for the kinematic model of the continuum robot: (a) a single segment of the 1-DoF structure with continuous backbones in body sections and (b) a single segment of the 2-DoF structure with twin-pivot compliant joints in tip sections.

Furthermore, as each section includes identical segments and two different section structures are applied in body and tip respectively, the configuration-cable kinematics can be expressed by analysing two single segments in which one is from the body (Fig. 3(a)) and the other is from tip (Fig. 3(b)).

**Body segment -** In the body sections (i.e. $i = 1,…,10$), a single segment, which includes two adjacent disks and a continuous compliant joint, represents a basic 1-DoF structure in $i^{th}$ section, as shown in Fig. 3(a). Noting that the through-disk cable lengths are constant, the contribution of arc parameters to the entire cable length only relates to the sum of changes of gap cable length $\Delta l_{i,j,k}$. According to the geometry relationships, the change of single gap cable length can be expressed as

$$\begin{cases} \Delta l_{i,j,k} = 2 \cdot \left|D_1 \cdot \sin\varphi_{i,j,k}\right| \cdot \left[\sin(\theta_{s1} - \frac{\theta_{i,j}}{2}) - \sin(\theta_{s1})\right]/\cos\theta_{s1} \\ \Delta l_{i,j,k+1} = 2 \cdot \left|D_1 \cdot \sin\varphi_{i,j,k+1}\right| \cdot \left[\sin(\theta_{s2} + \frac{\theta_{i,j}}{2}) - \sin(\theta_{s2})\right]/\cos\theta_{s2} \end{cases} \quad (3)$$

where $\theta_{s1}$ and $\theta_{s2}$ are the slope angles of bevels I and II, respectively, and $\theta_{i,j}$ is the segment bending angle. In this way, by considering all of the segments, the total change in length of the cables in the $i^{th}$ section (body) can be obtained.

**Tip segment -** In the tip sections (i.e. $i = 11,12,13$), a single segment, which includes three adjacent disks and two twin-pivot joints, represents a basic 2-DoF structure in the tip, as shown in Fig. 3(b). A complete derivation of the inverse kinematics of a twin-pivot compliant joint structure is given in [20]. Due to four (two paired) cables are used to shape the arc in [20], while in this work only a group of three cables through the bevelled disks are used. For this reason, the kinematics of the single segment needs to be discussed in detail. Referring to Fig. 3(b), the bending angles of twin-pivot joints $j$ and $j + 1$ (i.e. $\theta_{i,j}$ and $\theta_{i,j+1}$) can be expressed with respect to the bending and direction angles of the corresponding section. Based on the obtained angle values of $\theta_{i,j}$ and $\theta_{i,j+1}$ and similar geometry relationships in the body segment, the change of cable lengths in the gap between $j^{th}$ and $(j-1)^{th}$ disks can be expressed as

$$\begin{cases} \Delta l_{i,j,k-1} = 2 \cdot \left|D_3 \cdot \sin\varphi_{i,j,k-1}\right| \cdot \left[\sin(\theta_{s3} - \frac{\theta_{i,j}}{2}) - \sin(\theta_{s3})\right]/\cos\theta_{s3} \\ \Delta l_{i,j,k} = 2 \cdot \left|D_3 \cdot \sin\varphi_{i,j,k}\right| \cdot \left[\sin(\theta_{s3} - \frac{\theta_{i,j}}{2}) - \sin(\theta_{s3})\right]/\cos\theta_{s3} \\ \Delta l_{i,j,k+1} = 2 \cdot \left|D_3 \cdot \sin\varphi_{i,j,k+1}\right| \cdot \left[\sin(\theta_{s3} + \frac{\theta_{i,j}}{2}) - \sin(\theta_{s3})\right]/\cos\theta_{s3} \end{cases} \quad (4)$$

Similarly, the change of cable lengths in the gap between $j^{th}$ and $(j+1)^{th}$ disks can be written as

$$\begin{cases} \Delta l_{i,j+1,k-1} = 2 \cdot \left|D_3 \cdot \sin\varphi_{i,j+1,k-1}\right| \cdot \left[\sin(\theta_{s3} - \frac{\theta_{i,j+1}}{2}) - \sin(\theta_{s3})\right]/\cos\theta_{s3} \\ \Delta l_{i,j+1,k} = 2 \cdot \left|D_3 \cdot \sin\varphi_{i,j+1,k}\right| \cdot \left[\sin(\theta_{s3} + \frac{\theta_{i,j+1}}{2}) - \sin(\theta_{s3})\right]/\cos\theta_{s3} \\ \Delta l_{i,j+1,k+1} = 2 \cdot \left|D_3 \cdot \sin\varphi_{i,j+1,k+1}\right| \cdot \left[\sin(\theta_{s3} + \frac{\theta_{i,j+1}}{2}) - \sin(\theta_{s3})\right]/\cos\theta_{s3} \end{cases} \quad (5)$$

Therefore, according to (4) and (5), the total length changes the cables in the $i^{th}$ section (tip) can be also obtained by taking into account the disk number in each tip section.

*2) Configuration-Task Kinematics:* The purpose of configuration-task kinematic analysis is to solve the mathematics to describe the global position and orientation of the end-effector, i.e. $O_{14} = [x_{14}, y_{14}, z_{14}]^T$ and ${}^O R_{O_{14}}$ based on the configuration of all sections.

According to the above definition and assumption $A1$, the following transformations are made step by step with the assumption that a moving coordinate system $\{O_{mov}\}$ coincides with $\{O_{i,j-1}\}$ at the beginning.

**Body segment -** In the body sections (i.e. $i = 1,…,10$), taking a single 1-DoF segment for example, as shown in Fig. 3(a), $\{O_{mov}\}$ rotates around $x_{i,j-1}$-axis by an angle of $\theta_{i,j}$ and translates to $O_{i,j}$ at the same time, then $\{O_{mov}\}$ becomes coincident with $\{O_{i,j}\}$. With these coordinate transformations, the homogeneous transformation matrix from $\{O_{i,j}\}$ to $\{O_{i,j-1}\}$ can be written as

$$^{O_{i,j-1}}\mathbf{T}_{O_{i,j}} = \begin{bmatrix} \mathbf{R}_{x_{i,j-1}}(\theta_{i,j}) & \mathbf{P}_{i,j} \\ 0 & 1 \end{bmatrix} \quad (6)$$

where $\mathbf{R}_{x_{i,j-1}}(\theta_{i,j})$ is the rotation matrix around $x_{i,j-1}$-axis and $\mathbf{P}_{i,j}$ is the translation vector. Since each segment is assumed to bend with a constant curvature and arc length of $\Delta\ell$, so according to the (1) and the geometrical relationship, the translation vector $\mathbf{P}_{i,j}$ can be calculated as

$$\mathbf{P}_{i,j} = [0 \quad (1-\cos\theta_{i,j})\cdot \Delta\ell/\theta_{i,j} \quad \sin\theta_{i,j}\cdot \Delta\ell/\theta_{i,j}]^T \quad (7)$$

**Tip segment -** In the tip sections (i.e. $i = 11, 12, 13$), similarly taking a single 2-DoF segment for example, as shown in Fig. 3(b). Firstly, $\{O_{mov}\}$ rotates around $y_{i,j-1}$-axis by an angle of $\theta_{i,j}$ and simultaneously translates to $O_{i,j}$, then becomes coincident with $\{O_{i,j}\}$. Secondly, $\{O_{mov}\}$ rotates around $x_{i,j}$-axis by an angle of $\theta_{i,j+1}$ and translates to $O_{i,j+1}$ at the same time, then $\{O_{mov}\}$ becomes coincident with $\{O_{i,j+1}\}$. With these coordinate transformations, the homogeneous transformation matrix from $\{O_{i,j+1}\}$ to $\{O_{i,j-1}\}$ can be written as

$$^{O_{i,j-1}}\mathbf{T}_{O_{i,j+1}} = \begin{bmatrix} \mathbf{R}_{y_{i,j-1}}(\theta_{i,j}) & \mathbf{P}_{i,j} \\ 0 & 1 \end{bmatrix} \cdot \begin{bmatrix} \mathbf{R}_{x_{i,j}}(\theta_{i,j+1}) & \mathbf{P}_{i,j+1} \\ 0 & 1 \end{bmatrix} \quad (8)$$

Similarly, the translation vectors $\mathbf{P}_{i,j}$ and $\mathbf{P}_{i,j+1}$ can be calculated as

$$\begin{aligned} \mathbf{P}_{i,j} &= [(1-\cos\theta_{i,j})\cdot \Delta\ell/\theta_{i,j} \quad 0 \quad \sin\theta_{i,j}\cdot \Delta\ell/\theta_{i,j}]^T \\ \mathbf{P}_{i,j+1} &= [0 \quad (1-\cos\theta_{i,j+1})\cdot \Delta\ell/\theta_{i,j+1} \quad \sin\theta_{i,j+1}\cdot \Delta\ell/\theta_{i,j+1}]^T \end{aligned} \quad (9)$$

By taking all of the segments in both body and tip sections, the homogeneous transformation matrix of the entire continuum robot (i.e. from the end disk of $13^{th}$ section to the base disk of $1^{st}$ section) can be calculated as

$$^{O_{1,1}}\mathbf{T}_{O_{13,10}} = \prod_{i=1}^{10}\prod_{j=1}^{8} {}^{O_{i,j}}\mathbf{T}_{O_{i,j+1}} \cdot \prod_{i=11}^{13}\prod_{j=1}^{10} {}^{O_{i,j}}\mathbf{T}_{O_{i,j+1}} \quad (10)$$

Furthermore, once the continuum robot is setup in the working space with the end-effector installed at the end disk of $13^{th}$ section, the transformation matrices between $O$ and $O_{1,1}$ and between $O_{14}$ and $O_{13,10}$ can be defined as

$$\begin{aligned} ^{O}\mathbf{T}_{O_{1,1}} &= \begin{bmatrix} \mathbf{R}_X(\alpha) & \mathbf{R}_Y(\beta) & \mathbf{R}_Z(\gamma) & \mathbf{P}_O \\ 0 & 0 & 0 & 1 \end{bmatrix} \\ ^{O_{13,10}}\mathbf{T}_{O_{14}} &= \begin{bmatrix} \mathbf{R}_{x_{13,10}}(\alpha') & \mathbf{R}_{y_{13,10}}(\beta') & \mathbf{R}_{z_{13,10}}(\gamma') & \mathbf{P}_{O_{14}} \\ 0 & 0 & 0 & 1 \end{bmatrix} \end{aligned} \quad (11)$$

where $\alpha, \beta, \gamma, \alpha', \beta'$ and $\gamma'$ are the rotation angles around axes $X, Y, Z, x_{13,10}, y_{13,10}$, and $z_{13,10}$, respectively, and $\mathbf{P}_O$ and $\mathbf{P}_{O_{14}}$ are the translation vectors.

The global position and orientation of the end-effector, i.e. $O_{14} = [x_{14}, y_{14}, z_{14}]^T$ and $^0\mathbf{R}_{14}$ can be obtained in the following transformation matrix

$$\begin{aligned} ^{O}\mathbf{T}_{O_{14}} &= {}^{O}\mathbf{T}_{O_{1,1}} \cdot {}^{O_{1,1}}\mathbf{T}_{O_{13,10}} \cdot {}^{O_{13,10}}\mathbf{T}_{O_{14}} \\ &= {}^{O}\mathbf{T}_{O_{1,1}} \cdot \prod_{i=1}^{10}\prod_{j=1}^{8} {}^{O_{i,j}}\mathbf{T}_{O_{i,j+1}} \cdot \prod_{i=11}^{13}\prod_{j=1}^{10} {}^{O_{i,j}}\mathbf{T}_{O_{i,j+1}} \cdot {}^{O_{13,10}}\mathbf{T}_{O_{14}} \end{aligned} \quad (12)$$

To calculate the configuration-cable and configuration-task kinematics derived above, the continuum robot configuration (i.e. $\theta_{i,j}$) needs to be determined first, which will be analysed based on static modelling in the following subsection.

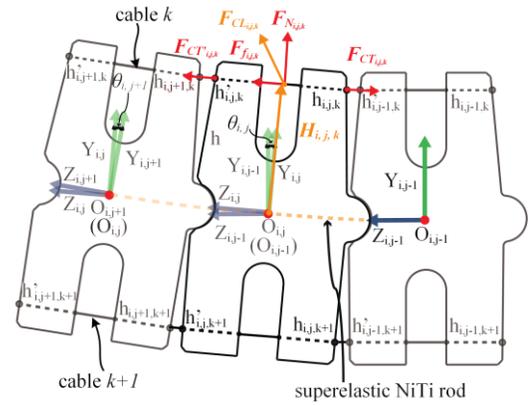

Fig. 4. Actuation cabling between two adjacent disks。

### D. Static Modelling

As a new structure of the continuum robot arm is presented in this paper by combining compliant and rigid joints, a new static modelling is proposed here by considering the applied forces and moments, which can be classified as four categories: actuation loading by the driving cables (i.e. tension, friction and contact force), gravity loading by the disks, elasticity loading by the backbones, and the external loading. Note that, based on a set of experiments, it was found that the friction coefficient varies with the change of the bending angle of a continuum robot section (detailed discussion in Section IV.A). Hence, the friction coefficient is considered as a non-constant parameter in this work, which is different from all the previous models.

**Actuation Loading -** The geometric relationship of the cable actuation loading applied to the $j^{th}$ disk of $i^{th}$ section is shown in Fig. 4. Within the local disk frame $\{O_{i,j-1}\}$, the cable actuation loading can be expressed by

$$\mathbf{F}_{CL_{i,j,k}}^{O_{i,j-1}} = \mathbf{F}_{N_{i,j,k}}^{O_{i,j-1}} + \mathbf{F}_{f_{i,j,k}}^{O_{i,j-1}} = \mathbf{F}_{CT'_{i,j,k}}^{O_{i,j-1}} + \mathbf{F}_{CT_{i,j,k}}^{O_{i,j-1}} \quad (13)$$

where the super- and subscripts respectively represent the frame and type of load, $\mathbf{F}_{N_{i,j,k}}^{O_{i,j-1}}$ is the normal force generated by the cable segment ($h'_{i,j,k}\,h_{i,j,k}$) against to the hole $H_{i,j,k}$, $\mathbf{F}_{f_{i,j,k}}^{O_{i,j-1}}$ is the friction between the $k^{th}$ cable and hole $H_{i,j,k}$, $\mathbf{F}_{CT'_{i,j,k}}^{O_{i,j-1}}$ and $\mathbf{F}_{CT_{i,j,k}}^{O_{i,j-1}}$ are respectively the cable tensions at the proximal position $h'_{i,j,k}$ and distal position $h_{i,j,k}$ of the hole $H_{i,j,k}$. Furthermore, the moment of the cable actuation loading $\mathbf{F}_{CL_{i,j,k}}^{O_{i,j-1}}$ relative to point $O_{i,j-1}$ can be obtained by vector cross product:

$$\mathbf{M}_{CL_{i,j,k}}^{O_{i,j-1}} = \mathbf{H}_{i,j,k}^{O_{i,j-1}} \times \mathbf{F}_{CL_{i,j,k}}^{O_{i,j-1}} \quad (14)$$

Since the driving cables follow a straight path between the routing holes in two adjacent disks, the direction of $F_{CT'_{i,j,k}}^{O_{i,j-1}}$, $F_{CT_{i,j,k}}^{O_{i,j-1}}$, $F_{N_{i,j,k}}^{O_{i,j-1}}$ and $F_{f_{i,j,k}}^{O_{i,j-1}}$ can be then determined by the coordinates of the hole $H_{i,j,k}$. By considering the transformation matrix mentioned in Section III.C, the separate segment cable loading and generated moment can be transformed back into the base disk of the $1^{st}$ section $O_1$.

In addition, as mentioned in the assumption $A_2$ at III.B.2), the frictional force can be calculated based on the Coulomb friction model and expressed as

$$\mathbf{F}_{f_{i,j,k}}^{O_{i,j-1}} = \mu_{\theta_{i,j}} \cdot \left|\mathbf{F}_{N_{i,j,k}}^{O_{i,j-1}}\right| = \mathbf{F}_{CT_{i,j,k}}^{O_{i,j-1}} - \mathbf{F}_{CT'_{i,j,k}}^{O_{i,j-1}} \quad (15)$$

where $\mu_{\theta_{i,j}}$ is the friction coefficient between the cable and the routing hole at the bending angle of $\theta_{i,j}$. Note that the friction coefficient in this work is a function of bending angle rather than a constant value and the detailed reason for this phenomenon will be discussed in Section IV.A.

**Gravity Loading** - Since the proposed continuum robot is with a highly slender design (diameter-to-length ration <0.02), the gravity effect of the supporting disks is non-negligible. As shown in Fig. 3, the gravity is defined along the negative direction of Z-axis of world frame $\{O\}$. Then for the $j^{th}$ disk of $i^{th}$ section, the homogeneous gravity vectors in $\{O\}$ can be expressed as:

$$\mathbf{G}_{i,j}^O = [0 \quad 0 \quad -m_{i,j} \cdot g \quad 0]^T \quad (16)$$

where $m_{i,j}$ is the mass of the $j^{th}$ disk of $i^{th}$ section and $g$ is the gravitational acceleration.

By considering the transformation matrix ${}^O\mathbf{T}_{O_{i,j-1}}$, the gravity force $\mathbf{G}_{i,j}^O$ and its moment relative to the point $O_{i,j-1}$ can be obtained in the local frame $\{O_{i,j-1}\}$ as

$$\mathbf{G}_{i,j}^{O_{i,j-1}} = ({}^O\mathbf{T}_{O_{i,j-1}})^{-1} \cdot \mathbf{G}_{i,j}^O \quad (17)$$

$$\mathbf{M}_{GL_{i,j}}^{O_{i,j-1}} = \mathbf{O}_{i,j}^{O_{i,j-1}} \times \mathbf{G}_{i,j}^{O_{i,j-1}} \quad (18)$$

**Elasticity Loading** - As mentioned in the design and assumption, the proposed continuum robot uses a pair or twin-pivot superelastic NiTi rods as the backbone which performs a pure planar bending in body sections and each tip segment. By applying the Kirchhoff elastic rod theory [21], the backbone deformation in each segment can be approximated as an arc. By considering a single segment, as shown in Fig. 2, the homogeneous vector of the bending moment of the backbone between $j^{th}$ and $(j-1)^{th}$ disks in the $i^{th}$ section can be obtained in the local frame $\{O_{i,j-1}\}$ as

$$\mathbf{M}_{EL_{i,j}}^{O_{i,j-1}} = [E_{i,j} \cdot I_{x_{i,j}} \cdot \Delta\ell/\theta_{i,j} \quad 0 \quad 0 \quad 0]^T \quad (19)$$

where $\Delta\ell/\theta_{i,j}$ is the bending curvature, $E_{i,j}$ is the Young's modulus and $I_{x_{i,j}}$ is the moment of inertia.

**External Loading** - Generally, the external loading is applied to the end-effector which is fixed on the last disk of the robot (i.e. $10^{th}$ disk of $13^{th}$ section). The external loading force and moment can be noted as $F_{EL}^O$ and $M_{EL}^O$, which are the equivalent lumped force and moment expressed in the world frame $\{O\}$. By taking into account of the homogeneous transformation matrix ${}^O\mathbf{T}_{O_{i,j}}$, the external loading force $F_{EL}^O$ and its moment $M_{EL}^O$ relative to the point $O_{i,j-1}$ can be obtained in the local frame $\{O_{i,j-1}\}$ as

$$\mathbf{F}_{EL}^{O_{i,j-1}} = ({}^O\mathbf{T}_{O_{i,j-1}})^{-1} \cdot \mathbf{F}_{EL}^O \quad (20)$$

$$\mathbf{M}_{EL}^{O_{i,j-1}} = ({}^O\mathbf{T}_{O_{i,j-1}})^{-1} \cdot \mathbf{M}_{EL}^O \quad (21)$$

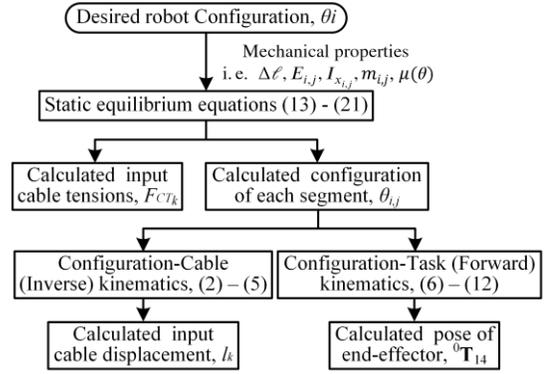

Fig. 5. Flowchart of kinematic and static analyses of the proposed modelling.

### E. Solutions

In Fig. 5, the detailed calculation process of the proposed modelling is presented with a flowchart. Note that the entire continuum robot shares the similar structure of the segments (i.e. joint units), the Newton-Euler formula is employed in this paper to establish the static equilibrium formulations of the whole robot, which can be obtained by recursion based on the above derived static equations (13)-(21) of any two adjacent units. Furthermore, with known the desired robot configuration ($\theta_i$) and mechanical properties (i.e. $\Delta\ell$, $E_{i,j}$, $I_{x_{i,j}}$, $m_{i,j}$ and $\mu(\theta)$), the input tensions and displacements for all of the actuation cables ($F_{CT_k}$ and $l_k$) can be obtained. In addition, the rotational angle of each segment ($\theta_{i,j}$) can be then calculated for predicting the profile of the continuum robot as well as the pose of end-effector (${}^O\mathbf{T}_{O_{14}}$).

## IV. EXPERIMENTAL VALIDATION

With the mechanical design and the key kinematic and static modelling commented, the validation tests (**i ~ iv**) of the capabilities of this complex mechatronic system has been performed to check some key characteristics to the targeted demonstration in this section. Firstly, a customised testbed (see Fig. 6(a)) was firstly built for: (**i**) measuring the variable friction coefficient, $\mu_\theta$, and analysing the relationship between $\mu_\theta$ and bending angle, $\theta$, (see Fig. 6(c)); (**ii**) validating the static modelling of a single body section without/with payload (see Figs. 7 and 8). Secondly, the entire continuum robot (see Fig. 9) was built for: (**iii**) testing the C-c curve following (see Fig. 10) and (**iv**) stiffness/payload at the distal end of the robot (Figs. 11 and 12).

### A. Static Model Validation and Analysis

**Test i** – In Fig. 6(a), an experimental set-up was built to obtain the variable friction coefficient $\mu_\theta$ in (15), where the identical body disk and stainless steel cable (Orminston®) from the continuum robot prototype were used. The driving cable passed through the routing hole of the disk which was fixed on `the frame with one end attached to the standard weight and the other end to a motor unit equipped with a load cell (Omega®). The angle $\theta$ between the driving cable and the routing hole was adjusted by a series of pulleys.

Based on this testbed, five sets of trials have been carried out by inputting five different standard weights (i.e. 0.8908 kg, 1.0908 kg, 2.0908 kg, 3.0908 kg and 4.0908 kg) at different angles between driving cable and the routing hole (i.e. $\theta = 1.25°$, $5°$, $10.25°$ and $20°$). Due to friction between the routing hole and the cable, the measured value of the load cell is always different from the standard weight attached at the other end of the driving cable and the difference can be obtained as the maximum static frictional force between the driving cable and the routing hole. Firstly, the experimental results are represented to determine the relationship (i.e. $\mu$) between the normal $F_N$, and the friction, $F_f$, forces at five different angles, as shown in Fig. 6(b), where measured values were plotted as discrete points and then represented by solid lines through the linear fitting. Therefore, the friction coefficient for different cable angles can be obtained as indicated in Fig. 6(b). Furthermore, the calculated values of $\mu$ at five different angles were plotted with the quadratic fitting line and the relationship between the friction coefficient and the angle $\theta$ was obtained,
as shown in Fig. 6(c). Note that all disks of the body sections in this work share the same function of the friction coefficient since they are made of the same material (NiTi-4Al).

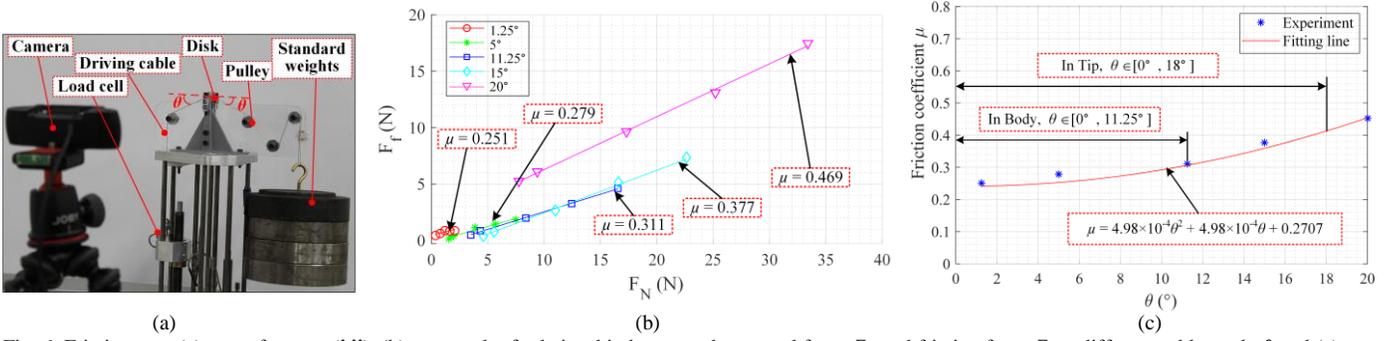

Fig. 6. Friction test: (a) setup for tests (**i,ii**), (b) test result of relationship between the normal force $F_N$ and friction force $F_f$ at different cable angle $\theta$ and (c) test result of friction coefficient vs cable angle $\theta$.

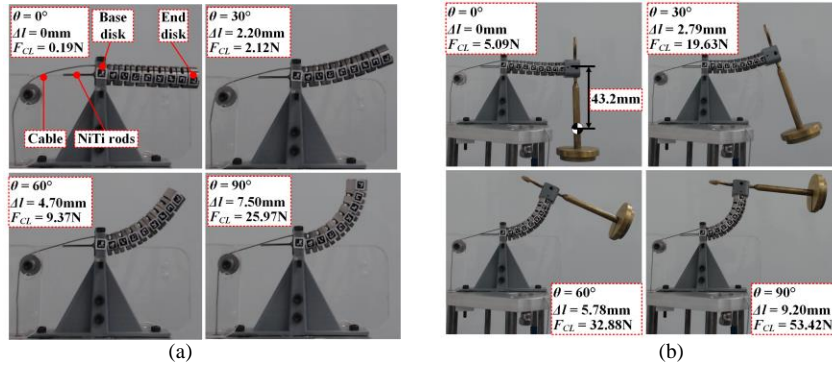

Fig. 7. Snapshots of single body section bending test for determing the relationship between desired and tested configurations : (a) without payload and (b) with payload of 90.8g.

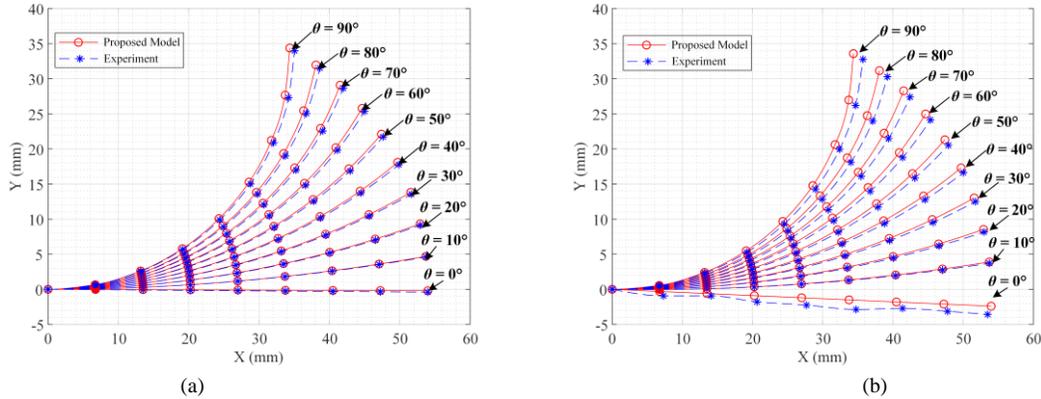

Fig. 8. Comparison between model-based (13) – (21) and experimental measurement positions of a body section: (a) without payload (average and maximum errors are 0.42 mm and 0.79 mm, accounting for 0.79% and 1.47% of overall section length) and (b) with payload of of 90.8g (average and maximum errors are 0.94 mm and 1.59 mm, accounting for 1.74% and 2.94% of the entire section length).

**Test ii** - A single section prototype was built to validate the proposed static model, (13) - (21) as presented in III.D. As shown in the first snapshot of Fig. 7(a), this prototype consists of 9 disks, where the first disk is fixed on the frame as the base and connected with the rest 8 disks (identical to one body section in the proposed continuum robot) with a pair of superelastic NiTi rods. A single cable is utilised for actuating this section with one end fixed at the tip disk and the other end connected to the motor unit with load cell and linear movement feedback. The section profile can be detected by a customized vision system consisting of a 1920 x 1080 pixel camera (Logitech®) and 9 Aruco makers. The camera was calibrated and initialised before experiments and the precision of the vision measuring system is $\pm$ 0.1mm [22]. It should be noted that this prototype only performs a planar movement and measurements are taken along the continuum robot backbone by detecting the coordinates of each disk centre.

Using this testbed, two groups of experiments are made to perform a planar bending ($\theta = 0°$ to $90°$) without and with a payload of 90.8g. Results are shown in Figs. 7 and 8, where test snapshots are presented with information of predefined bending angle (($\theta$), actuation cable displacement ($\Delta l$) and loading force ($F_{CL}$) (see Fig. 7) and experimental data are represented by blue stars and static-model-calculated data are represented by circles (see Fig. 8).

As shown in Figs. 7(a) and 8(a), the single section prototype performs very well the predefined bending curve under the gravity effect and the proposed static model has high accuracy in predicting the real prototype deflections. By comparing the experimental and model-based data, the average and the maximum errors are obtained as 0.43 mm and 0.79 mm, respectively, which only account for 0.79 % and 1.47 % of the entire length of the single section (i.e. 54 mm).

In Figs. 7(b) and 8(b), a standard weight of 90.8g is attached at the end disk of the single section prototype. As it can be seen, the built prototype can well perform the predefined bending movement and the simulated shape coincides well the real section shape. Further qualitative analysis shows that the average and maximum errors can be calculated as 0.94 mm and 1.59 mm with the payload, respectively, accounting for 1.74% and 2.94 % of the entire section length.

According to these tests, it can be found that the proposed kinematic and static models can be utilised for predicting the shape of the novel structure of the extra slender continuum robot when taking different payloads, with small errors. Hence, it was utilised when designing the mechanical structure of the robot arm.

*B. C-c curve Following Validation*

**Test iii** - To validate the capability of the proposed design solution performing the required C-c configuration shape at the specified curvature and centre (see Fig. 1), an experimental layout of the built prototype of the entire continuum robot is presented, as shown in Fig. 9. In the required static C-c shape, section 1 is forming the c-shape which indicates the navigation in/out at the port (see Fig. 1), while the other sections are forming the C-shape, which represents the navigation along the middle line of the combustor chamber (see Fig. 1).

The VICON ® optical motion capture system consisting of four high-resolution cameras was used to measure the movement with a standard set-up for capturing data. Five reflective markers have been added on each of the small square plates (15 x 15 mm $^2$), as shown in Fig. 9 with a detailed view. These plates are respectively mounted on the base of the continuum robot arm and the tip disks of each of the 13 sections, which allows the tracking of the position and orientation of each marker plate (i.e. representing the section frames $\{O_i\}$ and end-effector frame $\{O_{14}\}$). In addition, a low-level control system has been developed by using LabVIEW FPGA and load cells and linear encoders have been integrated with the actuation pack of each motor, allowing the close-loop control of cable lengths and tensions.

In Fig. 10, the experimental results are plotted to compare the actuated against the calculated positions of the continuum robot arm with two poses: the initial straight shape and the full C-c configuration shape. On both poses, the model-based calculated results are represented as a red solid line with circles whilst the VICON ® measured data are blue dashed lines with stars indicating the positions of markers.

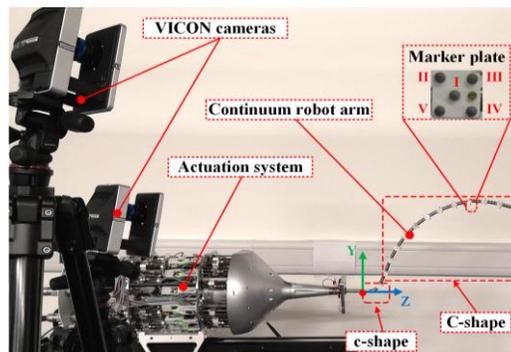

Fig. 9. An experimental layout of the continuum robot with VICON® system.

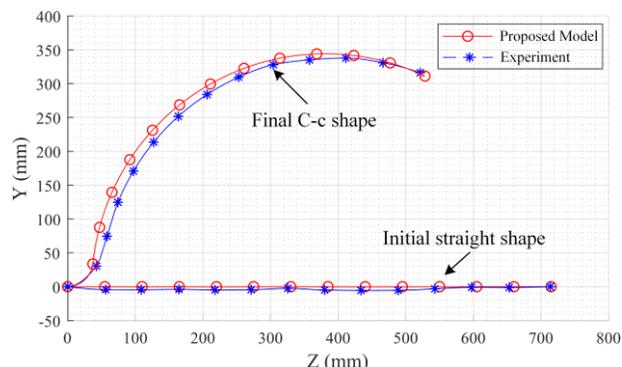

Fig. 10. A bending test to perform C-c configuration shape.

At the initial straight shape, there is a small error between the measured and simulated positions. The average and maximum errors are respectively calculated as 3.54 mm and 5.88 mm. Considering that the entire length of continuum robot arm is 715 mm, this error can be considered negligible as it only accounts for average of 0.49 % and maximum of 0.82 % as well as considering the assembly errors (position and orientation) of the marker plates on the continuum robot arm and the clearance between the holes and the cables.

At the final full C-c configuration shape, it can be noted that the positions of the entire shape of the continuum robot arm are slightly undershoot comparing with the simulated data., which is mainly caused by the cable tension reduction while passing through the cable guide. To evaluate the error in C-c shape, it can be characterised using two methods, by the resulting radius and the bending angle of the curve. The central point of the curve can be determined from where the normals of each marker plate intersect, while the curvature of each section can be calculated by the relation between adjacent markers. Furthermore, the experimental result of the radius can be calculated as 558.37 mm in Fig. 10 and the error is accounted for 1.02 % compared with the model-based one (564.14 mm). Note that the actuation of the cables mostly relies on the feedback from the linear encoders and load cells, the coupling of the actuation cables between different sections are not considered in this paper. Whilst the test results in this subsection demonstrate that the system is sufficient to perform the desired final C-c shape and to meet the requirements in static positioning.

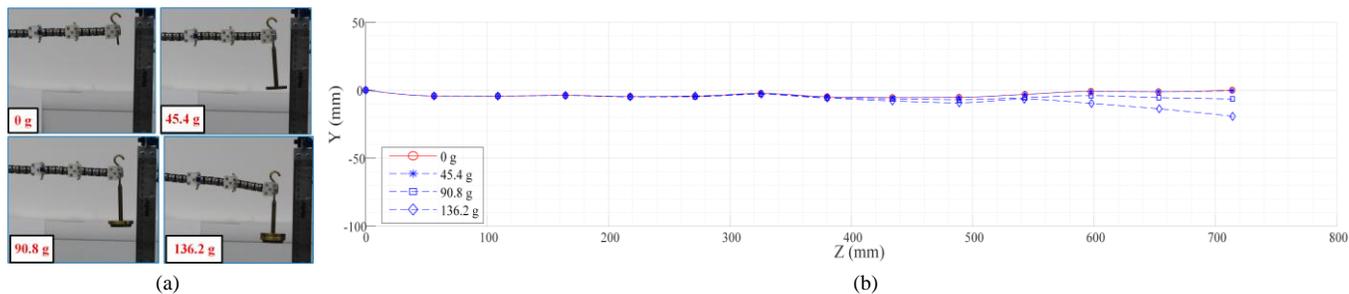

Fig. 11. Deflections of continuum robot caused by different weights in the straight shape: (a) snapshots and (b) test results

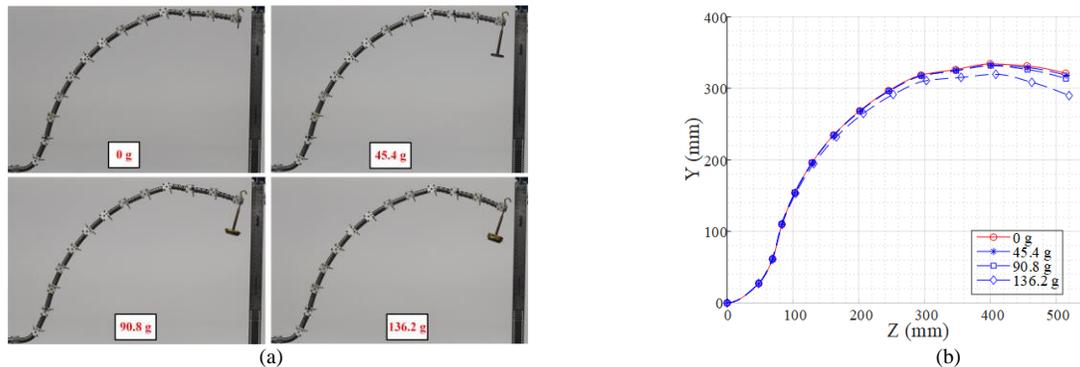

Fig. 12. Deflections of the continuum robot caused by different weights in the C-c shape: (a) snapshots and (b) test results.

Further work will focus on improving the kinematics performance of the extra slender robot arm on the dynamic positioning during the entire navigation process, by compensating the section coupling during the C-c shape transaction.

### C. Stiffness Capability Test

**Test iv** - Stiffness capability is crucial for the proposed continuum robot as it needs to carry appropriate end-effectors (e.g. inspection and repair tools) and perform precise in-situ operations in the aeroengine. To validate the stiffness capability, two groups of tests were carried out with the similar experimental set-up in Section IV.B by attaching different standard payloads (from 0 g to 136.2 g in increments of 45.4 g) in two typical configurations (i.e. straight and C-c shapes).

Experimental results are presented with a series of snapshots of the continuum robot prototype with an indication of the attached payloads (see Figs. 11(a) and 12(a)) and deflections are measured and plotted in Figs. 11(b) and 12(b). It is indicated that the continuum robot underwent slight deflections ($< 8$ mm, i.e. $< 1.12\%$ of the entire length of 715 mm) when payloads are less than 90.8 g. By increasing the payload to 136.2 g, the deflections are remaining under 20 mm (straight) and 32 mm (C-c shape) which are accounted for 2.79 % and 4.45 % of the entire length. It can be considered that the proposed continuum robot is capable of matching the payload requirement (i.e. the weight of the end-effector, 125 g) as mentioned in Section II.

### V. CONCLUSIONS

This paper presents an approach to design, model and validate a highly slender (diameter-to-length $<0.02$) continuum robot (16 DoFs) to be used for maintenance of aeroengine combustor via borescope ports, i.e. without the need to dismantle the enngine from aircrafts. To the best of authors' knowledge, the proposed solution is among the most slender and fully actuated continuum robot (diameter of 12.7 mm and length of 715 mm) with an appropriate stiffness that allows it to carry miniaturised industrial end-effectors (at least 125 g).

The novel mechanical design of the proposed continuum robot arm relies on a two-stage structure with 13 sections (ten 1-DoF body sections and three 2-DoF tip sections) based on customised bevelled disks and rigid-compliant joints. The piecewise constant-curvature approach and Kirchhoff elastic rod theory are utilised to comprehensively establish the kinematic and static models of the proposed continuum robot for the first time. It is worth noting that the kinematics and statics are based on segment scale and the friction coefficient is considered as a non-constant variable, approach which yields an imporeved performance of the rovot. Furthermore, kinematic and static models have been developed and configuration-cable kinematics has been solved to enable the control of the multi-section robot reaching the desired C-c configuration shape for feed in/out the combustor and further repair operation.

Finally, experimental validation is made with several built setups through four tests (i.e. friction, single section without/with payload, C-c shape following, stiffness in straight/C-c shape). It has been proved that: 1) the friction coefficient between the cable and routing hole is a function of bending angle rather than a constant value; 2) the proposed static model has high accuracy in predicting the in-plane bending shape of a single section, e.g. the average errors are 0.79 % and 1.74 % without/with payload of 90.8g; 3) the continuum robot is successfully built and tested by performing the desired C-c shape, e.g. the average deviation of curve radius is less than 1.02 %; 4) the continuum robot is capable of carrying a payload of 136.2g with the deflections under 2.79 % (straight) and 4.45 % (C-c shape). Thus, this paper presents an extra slender continuum robot that could be considered as a step forward to provide aeroengine manufacturers with a solution to perform complex maintenance (i.e. inspection and repair) tasks.

ACKNOWLEDGEMENT

The authors would like to thank David Palmer and David Alatorre for their assistance with the development of the proposed continuum robot system.